\documentclass[11pt,a4paper]{article}
\usepackage[hyperref]{naaclhlt2019}
\usepackage{times}
\usepackage{latexsym}

\usepackage{url}

\usepackage{natbib}
\usepackage{amsmath}
\usepackage{float}
\usepackage{caption}
\usepackage{subcaption}
\usepackage{threeparttable}
\usepackage{booktabs}
\usepackage{graphicx}
\usepackage{graphicx}
\usepackage{tabularx}

\usepackage{amssymb}

\newcommand{\spc}[2][c]{\begin{tabular}[#1]{@{}l@{}}#2\end{tabular}}

\usepackage{todonotes}

\usepackage{enumitem}

\DeclareGraphicsExtensions{.pdf,.png}
\graphicspath{{./figures/}}

\aclfinalcopy
\def\aclpaperid{2308}

\title{Learning Outside the Box: \\Discourse-level Features Improve Metaphor Identification}

\author{Jesse Mu{\normalfont \textsuperscript{1,3}},\,\, Helen Yannakoudakis{\normalfont \textsuperscript{2}},\,\, Ekaterina Shutova{\normalfont \textsuperscript{4}}\\
  \textsuperscript{1}Computer Science Department, Stanford University, USA\\
  \textsuperscript{2}The ALTA Institute, \textsuperscript{3}Dept. of CS \& Technology, University of Cambridge, UK\\
  \textsuperscript{4}ILLC, University of Amsterdam, The Netherlands\\
  \normalsize{{\tt muj@stanford.edu, hy260@cl.cam.ac.uk, e.shutova@uva.nl}}
}

\begin{document}

\maketitle

\begin{abstract}
Most current approaches to metaphor identification use restricted linguistic contexts, e.g.\ by
considering only a verb's arguments or the sentence containing a phrase. Inspired by pragmatic accounts of metaphor, we argue that broader discourse features are
crucial for better metaphor identification. We train simple gradient boosting classifiers on
representations of an utterance and its surrounding discourse learned with a variety of document
embedding methods, obtaining near state-of-the-art results on the 2018 VU Amsterdam metaphor
identification task without the complex metaphor-specific features or deep neural architectures
employed by other systems. A qualitative analysis further confirms the need for broader context in metaphor processing.
\end{abstract}

\section{Introduction}

From \emph{bottled up anger} to \emph{the world is your oyster}, metaphor is a defining component
of language, adding poetry and humor to communication
\citep{glucksberg2001understanding} and serving as a tool for reasoning about relations between
concepts \citep{lakoff1980metaphors}.
Designing metaphor processing systems
has thus seen significant interest in the NLP community, with applications from
information retrieval \citep{korkontzelos2013semeval} to machine translation
\citep{saygin2001processing}.

An important first step in any metaphor processing pipeline is metaphor \emph{identification}. To
date, most approaches to its identification operate in restricted contexts, for instance, by only
considering isolated verb--argument pairs (e.g.\ \emph{deflate economy})
\citep{rei2017grasping} or the sentence containing an utterance
\citep{gao2018neural}.
However, wider context is crucial for understanding metaphor: for instance, the phrase \emph{drowning students} can be interpreted as literal (in the context of \emph{water}) or metaphorical (in the context of \emph{homework}).
Often the context required extends beyond
the immediate sentence; in Table~\ref{tab:context_helps},
coreferences (\emph{them}) must be resolved to understand the arguments of a
verb, and a \emph{game} is metaphorical in a political context. Indeed, a rich
linguistic tradition \citep{grice1975logic,searle1979metaphor,sperber1986relevance} explains
metaphor as arising from violations of expectations in a conversational context.

Following these theories, in this paper we argue that metaphor processing models
should expand beyond restricted contexts to use representations of wider
discourse. We support this claim with two contributions: (1) we develop metaphor
identification models which take as input an utterance, its immediate lexico--syntactic context, and broader discourse
representations, and demonstrate that incorporating discourse features improves performance;  (2) we perform a qualitative analysis
 and show that broader context is often required to correctly interpret metaphors.
To the best of our knowledge, this is the first work to investigate the effects of broader 
discourse on metaphor identification.\footnote{Code and data available at \url{https://github.com/jayelm/broader-metaphor}.}
  
\begin{table}[t]
  \centering
  {
  \begin{tabular}{l}
    \toprule
    \spc{%
      \textcolor{gray}{``You can't steal their ideas.''} ``No, \\
      idiot---not so \underline{I} can \textbf{steal} \underline{them}.''
    } \\
    \midrule
    \spc{%
      Britain still can't decide when to \textbf{play} the \\
      mandarinate \underline{game} of of silence [\dots] \textcolor{gray}{interests} \\
      \textcolor{gray}{and concern of the Chinese government.}
    } \\
    \bottomrule
  \end{tabular}
  }
  \caption{Metaphorical examples from the VUA dataset in context. Verb is bolded, arguments underlined.
           Immediate sentence in black, with further context in gray.}
  \label{tab:context_helps}
  \vspace{-1em}
\end{table}

\section{Related work}

Metaphor identification is typically framed as a binary classification task, either with (1) word
tuples such as SVO triples (car \emph{drinks} gasoline) or (2) whole sentences as input, where the goal is
to predict the metaphoricity of a token in the sentence. Recent work has used a variety of
features extracted from these two types of contexts, including selectional preferences
\citep{shutova2013metaphor,beigmanklebanov2016semantic}, concreteness/imageability
\citep{turney2011literal,tsvetkov2014metaphor},
 multi-modal
\citep{tekiroglu2015exploring,shutova2016black} and neural features
\citep{do2016token,rei2017grasping}.

At the recent VU Amsterdam (VUA) metaphor identification shared task
\citep{leong2018report}, neural approaches dominated, with most teams using LSTMs
trained on word embeddings and additional linguistic features, such as semantic classes and part of
speech tags \citep{wu2018thungn,stemle2018using,mykowiecka2018detecting,swarnkar2018dilstm}. Most
recently, \citet{gao2018neural} revisited this task, reporting state-of-the-art results with BiLSTMs and contextualized word embeddings \citep{peters2018deep}.
 To the best of our knowledge, none of the existing approaches have utilized information from wider discourse context in metaphor identification, nor investigated its effects.

\section{Data}

Following past work, we use the \emph{Verbs} subset of the VUA metaphor corpus \citep{steen2010method}
used in the above shared task.  The data consists of $17240$ training and $5873$ test examples,
equally distributed across 4 genres of the British National Corpus: Academic, Conversation, News,
and Fiction. All verbs are annotated as metaphorical or literal in these texts. We sample $500$ examples randomly from the training set as a development set.

\section{Models}

For each utterance, our models learn generic representations of a \emph{verb lemma},\footnote{The lemmatized form of the verb has improved generalization in other systems \citep{beigmanklebanov2016semantic}.} its syntactic \emph{arguments}, and
its broader discourse \emph{context}. We concatenate these features into a single feature vector and feed them into a gradient boosting decision tree classifier \citep{chen2016xgboost}.\footnote{We use the default parameters of the XGBoost package: a maximum tree depth of 3, 100 trees, and $\eta = 0.1$.}
By observing performance
differences when using the lemma only (L), lemma + arguments (LA), or lemma + arguments + context
(LAC), we can investigate the effects of including broader context.

\begin{figure*}[t]
  \centering
  \includegraphics[width=0.8\linewidth]{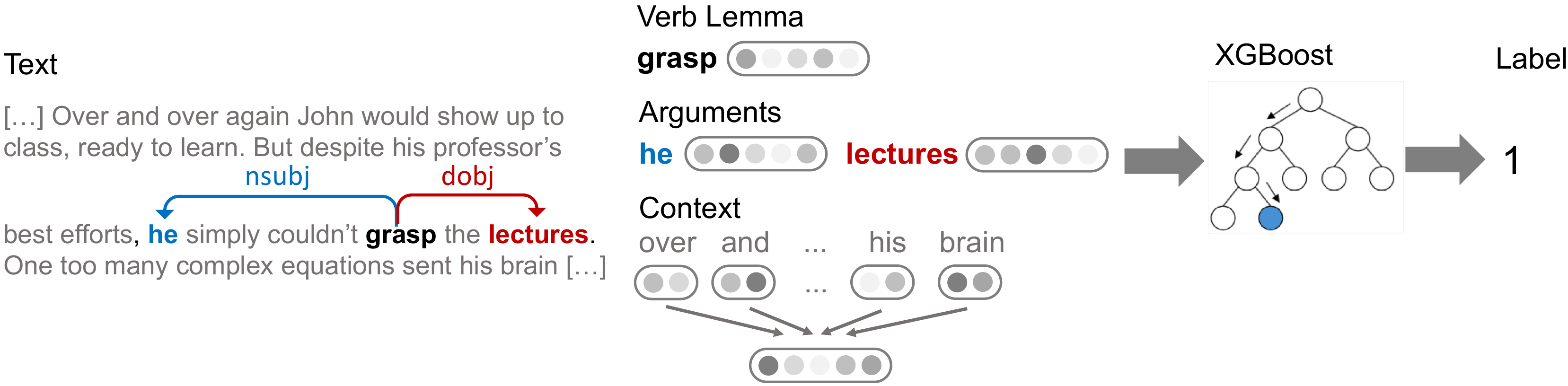}
  \caption{The general feature extraction and classification pipeline of our approach.}
  \label{fig:model_pipeline}
  \vspace{-1em}
\end{figure*}

To obtain arguments for verbs, we extract subjects and direct objects with Stanford CoreNLP
\citep{manning2014stanford}. $67.4\%$ of verb usages in the dataset have at least one argument;
absent arguments are represented as zero vectors. To obtain the broader \emph{context} of a
verb, we take its surrounding paragraph as defined by the BNC; the average
number of tokens in a context is $97.3$.  Figure~\ref{fig:model_pipeline} depicts the feature extraction and classification pipeline of our approach.

To learn representations, we use several widely-used embedding methods:\footnote{These methods differ significantly in dimensionality and training data. Our intent is not to exhaustively compare these methods, but rather claim generally that many embeddings give good performance on this task.}

\paragraph{GloVe} We use $300$-dimensional pre-trained GloVe embeddings \citep{pennington2014glove} trained on the Common Crawl corpus
as representations of a lemma and its arguments. To learn a context embedding, we simply average
the vectors of the tokens in the context. Out-of-vocabulary words are represented as a mean across all vectors.

\paragraph{doc2vec} We use pretrained $300$-dimensional paragraph vectors learned with the distributed bag-of-words method of
\citet{le2014distributed} (colloquially, doc2vec), trained on Wikipedia
\citep{lau2016empirical}. Here, paragraph vectors are learned to predict randomly sampled words
from the paragraph, ignoring word order. To extract representations for verbs and arguments, we
embed one-word ``documents'' consisting of only the word itself.\footnote{Since some methods provide only document embeddings and not word embeddings, for consistency, in all methods we use the same embedding process even for single-word verbs and arguments.}
We use a learning rate $\alpha = 0.01$ and $1000$ epochs to infer vectors.

\paragraph{Skip-thought} We use pretrained skip-thought vectors \citep{kiros2015skip} learned from training an encoder--decoder model to
reconstruct the surrounding sentences of an input sentence from the Toronto BooksCorpus  \citep{zhu2015aligning}. From this model, we extract $4800$-dimensional representations for verb lemma, arguments, and contexts.

\paragraph{ELMo} Finally, we use ELMo, a model of deep contextualized word embeddings \citep{peters2018deep}. We extract $1024$-dimensional representations from
the last layer of a stacked BiLSTM trained on Wikipedia and monolingual news data from WMT 2008--2012.
To learn
embeddings for verbs and arguments, we extract representations for sentences containing only the
word itself. To learn context embeddings, we again average the constituent word embeddings. 

\section{Evaluation}

For each embedding method, we evaluate the three configurations of features---L, LA, and LAC---on the VUA shared task train/test split,
reporting precision, recall and F1 score. Since we are interested in whether incorporating broader
context significantly improves identification performance, we compare successive model predictions
(LAC vs.\ LA; LA vs.\ L) using the \emph{mid-p} variant of McNemar's test for
paired binary data \citep{fagerland2013mcnemar}.

\subsection{Comparison Systems}
\label{ssec:context_baselines}

We first compare our models to the baselines of the VUA shared task \citep{leong2018report}:
\emph{Baseline 1}, a logistic regression classifier trained only on one-hot encodings of verb
lemmas; and \emph{Baseline 2}, the same classifier with additional WordNet class and concreteness
features. We also compare to the best systems submitted to the VUA shared task: \citet{wu2018thungn}, an ensemble of $20$ CNN-BiLSTMs trained on word2vec
embeddings, part-of-speech tags, and word embedding clusters; and \citet{stemle2018using}, a BiLSTM
trained on embeddings from English language learner corpora.

\begin{table}[t]
  \centering
  { \small
    \begin{threeparttable}
  \begin{tabular}{lllll}
    \toprule
    \multicolumn{2}{l}{Model} & P & R & F1 \\
    \midrule
    \multicolumn{2}{l}{Baseline 1 (lemma)}  & 51.0 & 65.4 & 57.3 \\
    \multicolumn{2}{l}{Baseline 2 (+WN, concrete)}  & 52.7 & 69.8 & 60.0 \\
    \multicolumn{2}{l}{\citet{stemle2018using}} & 54.7 & \textbf{77.9} & 64.2 \\
    \multicolumn{2}{l}{\citet{wu2018thungn}} & \textbf{60.0} & 76.3 & \textbf{67.2} \\
    \midrule
    GloVe & L (lemma) & 51.6 & 74.1 & 60.8 \\
          & LA (+ args)& 54.0 & 74.4 & 62.6\textsuperscript{***} \\
          & LAC (+ ctx)& 56.7 & 76.8 & 65.2\textsuperscript{***} \\
    \midrule
    doc2vec & L & 48.8 & 72.1 & 58.2 \\
          & LA & 50.5 & 71.4 & 59.1\textsuperscript{**} \\
        & LAC & 52.7 & 72.2 & 60.9\textsuperscript{***} \\
    \midrule
    skip-thought & L & 53.5 & 76.1 & 62.8 \\
                 & LA & 57.0 & 74.0 & 64.3\textsuperscript{***} \\
                 & LAC & 59.5 & 75.4 & 66.5\textsuperscript{***} \\
    \midrule
    ELMo & L & 51.3 & 74.9 & 60.9 \\
         & LA & 56.0 & 73.5 & 63.6\textsuperscript{***} \\
         & LAC & 58.9 & 77.1 & 66.8\textsuperscript{***} \\
    \bottomrule
  \end{tabular}
  \begin{tablenotes}
  \item[**,***]Significant improvement over previous model ($p < 0.01, 0.001$).
  \end{tablenotes}
  \end{threeparttable}
  }
   \caption{Metaphor identification results.}
  \label{tab:identification_context_results}
  \vspace{-1em}
\end{table}

\begin{table}[t]
  \centering
  {\small
  \begin{tabular}{llrrr}
    \toprule
    Genre & Model & P & R & F1 \\
    \midrule
    Academic   & Baseline 2 & 70.7 & 83.6 & \textbf{76.6} \\
               & \citet{wu2018thungn} & \textbf{74.6} & 76.3 & 75.5 \\
    & ELMo LAC & 65.4 & \textbf{86.8} & 74.6 \\
    \midrule
    Conversation   & Baseline 2 & 30.1 & 82.1 & 44.1 \\
                   & \citet{wu2018thungn} & 40.3 & 65.6 & \textbf{50.3} \\
    & ELMo LAC & \textbf{42.6} & 56.0 & 48.4 \\
    \midrule
    Fiction   & Baseline 2 & 40.7 & 66.7 & 50.6 \\
              & \citet{wu2018thungn} & \textbf{54.5} & \textbf{78.4} & \textbf{57.6} \\
    & ELMo LAC & 48.2 & 63.0 & 54.6 \\
    \midrule
    News      & Baseline 2 & 67.7 & 68.9 & 68.3 \\
              & \citet{wu2018thungn} & \textbf{69.4} & 74.4 & 71.8 \\
              & ELMo LAC & 65.2 & \textbf{80.0} & \textbf{71.8} \\
    \bottomrule
  \end{tabular}
  }
   \caption{Performance breakdown by genre for ELMo LAC model and comparison systems.}
  \label{tab:breakdown}
  \vspace{-0.5em}
\end{table}

\begin{table}[t]
  \centering
  {
    \begin{threeparttable}
  \begin{tabular}{lrrr}
    \toprule
          & Args & Sentence & Paragraph \\
    \midrule
    Overall & 40 & 49 & 11  \\
    \midrule
    \multicolumn{4}{l}{Model errors} \\
    ELMo L     & 37   & 50 & 13 \\
    ELMo LA     & 36   & 49 & 15 \\
    ELMo LAC     & 39   & 53 & 8 \\
    \bottomrule
  \end{tabular}
  \end{threeparttable}
  }
  \caption{Types of context required to interpret metaphors in the development set, both overall (first row) and for model errors. Each row is a separate (but overlapping) sample from the development set.}
  \label{tab:context_examples}
  \vspace{-1em}
\end{table}

\begin{table*}[t]
  \centering
  \begin{tabular}{llll}
    \toprule
    Sentence & Gold label & LA & LAC \\
    \midrule
    \spc{
      \textcolor{gray}{A major complication [\dots] is that the environment can rarely be treated} \\
      \textcolor{gray}{as in a laboratory experiment.} \textbf{Given} \underline{this}, determining the nature of the \\
      interactions between the variables becomes a matter of major difficulty.
    }
    & 0 & 1 & 0 \\
    \midrule
    \spc{
      \textcolor{gray}{For example, on high policy common opinion said that there was nothing} \\
      \textcolor{gray}{for it but to stay in the ERM.} \underline{He} \textbf{stayed} in, and the recession worsened. 
    }
    & 1 & 0 & 1 \\
    \bottomrule
  \end{tabular}
  \caption{Examples where context helps, along with gold label (0 -- literal; 1 -- metaphor) and model predictions (LA, LAC). Verb is
  bolded, arguments underlined. Additional context (needed for interpretation) in gray.}
  \vspace{-1em}
  \label{tab:la_wrong_lac_right}
\end{table*}

\subsection{Results}

Results for our models are presented in Table~\ref{tab:identification_context_results}.
Interestingly, most of the
simple lemma models (L) already perform at Baseline 2 level, obtaining F1 scores in the range $60$--$62$. This is likely due to the generalization made possible by dense representations of lemmas (vs. one-hot encodings) and the more powerful statistical classifier used.
As expected, the addition of argument information consistently enhances performance.

Crucially, the addition of broader discourse context improves performance for all embedding methods. In general, we observe
consistent, statistically significant increases of $2$-$3$ F1 points for incorporating discourse. Overall, all LAC models except doc2vec exhibit high performance, and would have achieved second place in the VUA shared task. These results show a clear trend: the incorporation of discourse information leads to improvement of metaphor identification performance across models. 

Table 3 displays the performance breakdown by genre in the VUA test set for our best performing model (ELMo LAC) and selected comparison systems. Echoing \citet{leong2018report}, we observe that the \textit{Conversation} and \textit{Fiction} genres are consistently more difficult than the \textit{Academic} and \textit{News} genres across all models. This is partially because in this dataset, metaphors in these genres are rarer, occuring $35\%$ of the time in \textit{Academic} and $43\%$ in \textit{News}, but only  
$15\%$ in \textit{Conversation} and $24\%$ in \textit{Fiction}. In addition, for our model specifically, Conversation genre contexts are much shorter on average ($23.8$ vs.\ $97.3$).

Our best performing model (ELMo LAC) is within $0.4$ F1 score of the first-place model in the VUA shared task \citep{wu2018thungn}. The GloVe LAC model would
also have obtained second place at $65.2$ F1, yet is considerably simpler than the systems used in the shared task, which employed ensembles of deep neural architectures and hand-engineered, metaphor-specific features.

\section{Qualitative analysis}

To better understand the ways in which discourse information plays a role in metaphor processing, we randomly sample $100$ examples from our
development set and manually categorize them by the amount of context required for their interpretation. For instance, a
verb may be interpretable when given just its arguments (direct subject/object), it may require context
from the enclosing sentence, or it may require paragraph-level context (or beyond). We also similarly analyze $100$
sampled errors made on the development set by the ELMo L, LA, and LAC models, to examine whether error types vary between models.

Our analysis in Table~\ref{tab:context_examples} shows that $11\%$ of examples in the development set require
paragraph-level context for correct interpretation. Indeed, while such examples are frequently misclassified by the L and LA models
($13\%$, $15\%$), the error rate is halved when context is included ($8\%$).

Table~\ref{tab:la_wrong_lac_right} further presents examples requiring at least paragraph-level context, along with gold label and model predictions. Out of the $31$ unique such examples identified in the above analyses, we found $11$ ($35\%$) requiring explicit coreference resolution of a pronoun or otherwise underspecified noun (e.g.\ Table~\ref{tab:la_wrong_lac_right} row 1) and $5$ ($16\%$) which reference an entity or event implicitly (\emph{ellipsis}; e.g.\ Table~\ref{tab:la_wrong_lac_right} row 2). However, we also observed $4$ errors ($13\%$) due to examples with non-verbs and incomplete sentences and $11$ examples ($35\%$) where not even paragraph-level context was sufficient for interpretation, mostly in the \textit{Conversation} genre, demonstrating the subjective and borderline nature of many of the annotations.

This analysis shows \emph{a priori} the need for broader context beyond sentence-level for robust metaphor processing. Yet this is not an upper bound on
performance gains; the general improvement of the LAC models over LA shows that
even when context is not strictly necessary, it can still be a useful signal for identification.

\section{Conclusion}

We presented the first models which leverage representations of discourse for metaphor identification.
The performance gains of these models demonstrate that incorporating broader discourse information is a
powerful feature for metaphor identification systems, aligning with our qualitative analysis and the theoretical and empirical evidence suggesting
metaphor comprehension is heavily influenced by wider context.

Given the
simplicity of our representations of context in these models, we are interested in future models
which (1) use discourse in more sophisticated ways, e.g.\ by modeling discourse relations or dialog state tracking \citep{henderson2015machine}, and (2) leverage more sophisticated neural architectures \citep{gao2018neural}.

\section*{Acknowledgments}

We thank anonymous reviewers for their insightful comments, Noah Goodman, and Ben Leong for assistance with the 2018 VUA shared task data. We thank the Department of Computer Science and Technology and Churchill College, University of Cambridge for travel funding. Jesse Mu is supported by a Churchill Scholarship and an NSF Graduate Research Fellowship. Helen Yannakoudakis was supported by Cambridge Assessment, University of Cambridge. We thank the NVIDIA Corporation for the donation of the Titan GPU used in this research.

\bibliography{metaphor_naacl19}

\begin{thebibliography}{32}
\expandafter\ifx\csname natexlab\endcsname\relax\def\natexlab#1{#1}\fi

\bibitem[{Beigman~Klebanov et~al.(2016)Beigman~Klebanov, Leong, Gutierrez,
  Shutova, and Flor}]{beigmanklebanov2016semantic}
Beata Beigman~Klebanov, Chee~Wee Leong, E.~Dario Gutierrez, Ekaterina Shutova,
  and Michael Flor. 2016.
\newblock Semantic classifications for detection of verb metaphors.
\newblock In \emph{Proceedings of the 54th Annual Meeting of the Association
  for Computational Linguistics (Volume 2: Short Papers)}, pages 101--106.

\bibitem[{Chen and Guestrin(2016)}]{chen2016xgboost}
Tianqi Chen and Carlos Guestrin. 2016.
\newblock {XGB}oost: A scalable tree boosting system.
\newblock In \emph{Proceedings of the 22nd ACM SIGKDD International Conference
  on Knowledge Discovery and Data Mining}, pages 785--794.

\bibitem[{Do~Dinh and Gurevych(2016)}]{do2016token}
Erik-L{\^a}n Do~Dinh and Iryna Gurevych. 2016.
\newblock Token-level metaphor detection using neural networks.
\newblock In \emph{Proceedings of the Fourth Workshop on Metaphor in NLP},
  pages 28--33.

\bibitem[{Fagerland et~al.(2013)Fagerland, Lydersen, and
  Laake}]{fagerland2013mcnemar}
Morten~W Fagerland, Stian Lydersen, and Petter Laake. 2013.
\newblock The {McNemar} test for binary matched-pairs data: mid-p and
  asymptotic are better than exact conditional.
\newblock \emph{BMC Medical Research Methodology}, 13(1):91.

\bibitem[{Gao et~al.(2018)Gao, Choi, Choi, and Zettlemoyer}]{gao2018neural}
Ge~Gao, Eunsol Choi, Yejin Choi, and Luke Zettlemoyer. 2018.
\newblock Neural metaphor detection in context.
\newblock In \emph{Proceedings of the 2018 Conference on Empirical Methods in
  Natural Language Processing}, pages 607--613.

\bibitem[{Glucksberg and McGlone(2001)}]{glucksberg2001understanding}
Sam Glucksberg and Matthew~S McGlone. 2001.
\newblock \emph{Understanding figurative language: From metaphor to idioms}.
\newblock Oxford University Press, Oxford.

\bibitem[{Grice(1975)}]{grice1975logic}
Herbert~P Grice. 1975.
\newblock Logic and conversation.
\newblock In Peter Cole and Jerry~L Morgan, editors, \emph{Syntax and
  Semantics}, volume~3, pages 41--58. Academic Press, New York.

\bibitem[{Henderson(2015)}]{henderson2015machine}
Matthew Henderson. 2015.
\newblock Machine learning for dialog state tracking: A review.
\newblock In \emph{Proceedings of The First International Workshop on Machine
  Learning in Spoken Language Processing}.

\bibitem[{Kiros et~al.(2015)Kiros, Zhu, Salakhutdinov, Zemel, Urtasun,
  Torralba, and Fidler}]{kiros2015skip}
Ryan Kiros, Yukun Zhu, Ruslan~R Salakhutdinov, Richard Zemel, Raquel Urtasun,
  Antonio Torralba, and Sanja Fidler. 2015.
\newblock Skip-thought vectors.
\newblock In \emph{Advances in Neural Information Processing Systems}, pages
  3294--3302.

\bibitem[{Korkontzelos et~al.(2013)Korkontzelos, Zesch, Zanzotto, and
  Biemann}]{korkontzelos2013semeval}
Ioannis Korkontzelos, Torsten Zesch, Fabio~Massimo Zanzotto, and Chris Biemann.
  2013.
\newblock Sem{E}val-2013 task 5: Evaluating phrasal semantics.
\newblock In \emph{Second Joint Conference on Lexical and Computational
  Semantics (* SEM), Volume 2: Proceedings of the Seventh International
  Workshop on Semantic Evaluation (SemEval 2013)}, volume~2, pages 39--47.

\bibitem[{Lakoff and Johnson(1980)}]{lakoff1980metaphors}
George Lakoff and Mark Johnson. 1980.
\newblock \emph{Metaphors We Live By}.
\newblock University of Chicago Press, Chicago.

\bibitem[{Lau and Baldwin(2016)}]{lau2016empirical}
Jey~Han Lau and Timothy Baldwin. 2016.
\newblock An empirical evaluation of doc2vec with practical insights into
  document embedding generation.
\newblock In \emph{Proceedings of the 1st Workshop on Representation Learning
  for NLP}, pages 78--86.

\bibitem[{Le and Mikolov(2014)}]{le2014distributed}
Quoc Le and Tomas Mikolov. 2014.
\newblock Distributed representations of sentences and documents.
\newblock In \emph{Proceedings of the 31st International Conference on Machine
  Learning}, pages 1188--1196.

\bibitem[{Leong et~al.(2018)Leong, Beigman~Klebanov, and
  Shutova}]{leong2018report}
Chee Wee~(Ben) Leong, Beata Beigman~Klebanov, and Ekaterina Shutova. 2018.
\newblock A report on the 2018 {VUA} metaphor detection shared task.
\newblock In \emph{Proceedings of the Workshop on Figurative Language
  Processing}, pages 56--66.

\bibitem[{Manning et~al.(2014)Manning, Surdeanu, Bauer, Finkel, Bethard, and
  McClosky}]{manning2014stanford}
Christopher Manning, Mihai Surdeanu, John Bauer, Jenny Finkel, Steven Bethard,
  and David McClosky. 2014.
\newblock The {Stanford CoreNLP} natural language processing toolkit.
\newblock In \emph{Proceedings of 52nd Annual Meeting of the Association for
  Computational Linguistics: System Demonstrations}, pages 55--60.

\bibitem[{Mykowiecka et~al.(2018)Mykowiecka, Wawer, and
  Marciniak}]{mykowiecka2018detecting}
Agnieszka Mykowiecka, Aleksander Wawer, and Malgorzata Marciniak. 2018.
\newblock Detecting figurative word occurrences using recurrent neural
  networks.
\newblock In \emph{Proceedings of the Workshop on Figurative Language
  Processing}, pages 124--127.

\bibitem[{Pennington et~al.(2014)Pennington, Socher, and
  Manning}]{pennington2014glove}
Jeffrey Pennington, Richard Socher, and Christopher Manning. 2014.
\newblock Glo{V}e: Global vectors for word representation.
\newblock In \emph{Proceedings of the 2014 Conference on Empirical Methods in
  Natural Language Processing}, pages 1532--1543.

\bibitem[{Peters et~al.(2018)Peters, Neumann, Iyyer, Gardner, Clark, Lee, and
  Zettlemoyer}]{peters2018deep}
Matthew Peters, Mark Neumann, Mohit Iyyer, Matt Gardner, Christopher Clark,
  Kenton Lee, and Luke Zettlemoyer. 2018.
\newblock Deep contextualized word representations.
\newblock In \emph{Proceedings of the 2018 Conference of the North American
  Chapter of the Association for Computational Linguistics: Human Language
  Technologies, Volume 1 (Long Papers)}, volume~1, pages 2227--2237.

\bibitem[{Rei et~al.(2017)Rei, Bulat, Kiela, and Shutova}]{rei2017grasping}
Marek Rei, Luana Bulat, Douwe Kiela, and Ekaterina Shutova. 2017.
\newblock Grasping the finer point: A supervised similarity network for
  metaphor detection.
\newblock In \emph{Proceedings of the 2017 Conference on Empirical Methods in
  Natural Language Processing}, pages 1537--1546.

\bibitem[{Saygin(2001)}]{saygin2001processing}
Ayse~Pinar Saygin. 2001.
\newblock Processing figurative language in a multi-lingual task: Translation,
  transfer and metaphor.
\newblock In \emph{Proceedings of the Workshop on Corpus-based and Processing
  Approaches to Figurative Language}.

\bibitem[{Searle(1979)}]{searle1979metaphor}
John Searle. 1979.
\newblock Metaphor.
\newblock In \emph{Expression and Meaning: Studies in the Theory of Speech
  Acts}, pages 76--116. Cambridge University Press, Cambridge and New York.

\bibitem[{Shutova(2013)}]{shutova2013metaphor}
Ekaterina Shutova. 2013.
\newblock Metaphor identification as interpretation.
\newblock In \emph{Second Joint Conference on Lexical and Computational
  Semantics (*SEM), Volume 1: Proceedings of the Main Conference and the Shared
  Task: Semantic Textual Similarity}, pages 276--285.

\bibitem[{Shutova et~al.(2016)Shutova, Kiela, and Maillard}]{shutova2016black}
Ekaterina Shutova, Douwe Kiela, and Jean Maillard. 2016.
\newblock Black holes and white rabbits: Metaphor identification with visual
  features.
\newblock In \emph{Proceedings of the 2016 Conference of the North American
  Chapter of the Association for Computational Linguistics: Human Language
  Technologies}, pages 160--170.

\bibitem[{Sperber and Wilson(1986)}]{sperber1986relevance}
Dan Sperber and Deirdre Wilson. 1986.
\newblock \emph{Relevance: Communication and cognition}.
\newblock Harvard University Press, Cambridge, MA.

\bibitem[{Steen et~al.(2010)Steen, Dorst, Herrmann, Kaal, Krennmayr, and
  Pasma}]{steen2010method}
Gerard~J Steen, Aletta~G Dorst, J~Berenike Herrmann, Anna Kaal, Tina Krennmayr,
  and Trijntje Pasma. 2010.
\newblock \emph{A method for linguistic metaphor identification: From {MIP} to
  {MIPVU}}.
\newblock John Benjamins Publishing Company, Amsterdam.

\bibitem[{Stemle and Onysko(2018)}]{stemle2018using}
Egon Stemle and Alexander Onysko. 2018.
\newblock Using language learner data for metaphor detection.
\newblock In \emph{Proceedings of the Workshop on Figurative Language
  Processing}, pages 133--138.

\bibitem[{Swarnkar and Singh(2018)}]{swarnkar2018dilstm}
Krishnkant Swarnkar and Anil~Kumar Singh. 2018.
\newblock {Di-LSTM} contrast : A deep neural network for metaphor detection.
\newblock In \emph{Proceedings of the Workshop on Figurative Language
  Processing}, pages 115--120.

\bibitem[{Tekiroglu et~al.(2015)Tekiroglu, {\"O}zbal, and
  Strapparava}]{tekiroglu2015exploring}
Serra~Sinem Tekiroglu, G{\"o}zde {\"O}zbal, and Carlo Strapparava. 2015.
\newblock Exploring sensorial features for metaphor identification.
\newblock In \emph{Proceedings of the Third Workshop on Metaphor in NLP}, pages
  31--39.

\bibitem[{Tsvetkov et~al.(2014)Tsvetkov, Boytsov, Gershman, Nyberg, and
  Dyer}]{tsvetkov2014metaphor}
Yulia Tsvetkov, Leonid Boytsov, Anatole Gershman, Eric Nyberg, and Chris Dyer.
  2014.
\newblock Metaphor detection with cross-lingual model transfer.
\newblock In \emph{Proceedings of the 52nd Annual Meeting of the Association
  for Computational Linguistics (Volume 1: Long Papers)}, volume~1, pages
  248--258.

\bibitem[{Turney et~al.(2011)Turney, Neuman, Assaf, and
  Cohen}]{turney2011literal}
Peter~D Turney, Yair Neuman, Dan Assaf, and Yohai Cohen. 2011.
\newblock Literal and metaphorical sense identification through concrete and
  abstract context.
\newblock In \emph{Proceedings of the Conference on Empirical Methods in
  Natural Language Processing}, pages 680--690.

\bibitem[{Wu et~al.(2018)Wu, Wu, Chen, Wu, Yuan, and Huang}]{wu2018thungn}
Chuhan Wu, Fangzhao Wu, Yubo Chen, Sixing Wu, Zhigang Yuan, and Yongfeng Huang.
  2018.
\newblock Neural metaphor detecting with {CNN-LSTM} model.
\newblock In \emph{Proceedings of the Workshop on Figurative Language
  Processing}, pages 110--114.

\bibitem[{Zhu et~al.(2015)Zhu, Kiros, Zemel, Salakhutdinov, Urtasun, Torralba,
  and Fidler}]{zhu2015aligning}
Yukun Zhu, Ryan Kiros, Rich Zemel, Ruslan Salakhutdinov, Raquel Urtasun,
  Antonio Torralba, and Sanja Fidler. 2015.
\newblock Aligning books and movies: Towards story-like visual explanations by
  watching movies and reading books.
\newblock In \emph{The IEEE International Conference on Computer Vision
  (ICCV)}, pages 19--27.

\end{thebibliography}
\bibliographystyle{acl_natbib}

\end{document}